\documentclass[11pt]{article}

\usepackage[preprint]{acl}

\usepackage{times}
\usepackage{latexsym}

\usepackage[T1]{fontenc}

\usepackage[utf8]{inputenc}

\usepackage{microtype}

\usepackage{inconsolata}

\usepackage{graphicx}

\usepackage{booktabs}
\usepackage{multirow}
\usepackage{array}
\usepackage{geometry}
\usepackage{colortbl}
\usepackage{xcolor}
\usepackage{stfloats} 
\usepackage{float} 
\usepackage{afterpage}

\title{CNSight: Evaluation of Clinical Note Segmentation Tools}

\author{
  Risha Surana, Adrian Law, Sunwoo Kim, Rishab Sridhar, Angxiao Han, Peiyu Hong  \\
  University of Southern California
}

\begin{document}
\maketitle
\begin{abstract}
Clinical notes are often stored in unstructured or semi-structured formats after extraction from electronic medical record (EMR) systems, which complicates their use for secondary analysis and downstream clinical applications. Reliable identification of section boundaries is a key step toward structuring these notes, as sections such as history of present illness, medications, and discharge instructions each provide distinct clinical contexts. In this work, we evaluate rule-based baselines, domain-specific transformer models, and large language models for clinical note segmentation using a curated dataset of 1,000 notes from MIMIC-IV. Our experiments show that large API-based models achieve the best overall performance, with GPT-5-mini reaching a best average F1 of 72.4 across sentence-level and freetext segmentation. Lightweight baselines remain competitive on structured sentence-level tasks but falter on unstructured freetext. Our results provide guidance for method selection and lay the groundwork for downstream tasks such as information extraction, cohort identification, and automated summarization.  

\end{abstract}

\section{Introduction}

EHR data is often processed and presented in plain text format for secondary use in modeling and data retrieval tasks. Clinical notes contain a wide range of information such as chief complaints, physician observations, and past medical history that can supplement structured data like lab values or medications. However the note text itself is often difficult to process with traditional or out-of-the-box models because of domain specific issues including medical abbreviations, semi-structured text, unrelated or redundant documentation, and information that varies depending on the patient or encounter.  

The first step in analyzing clinical notes is to identify sections from the full text and segment notes into distinct categories. Our goal in this work is to evaluate the note segmentation effectiveness of different models from open source libraries and prior research using a curated dataset.

\begin{figure*}[t]
    \centering
    \includegraphics[width=0.9\textwidth]{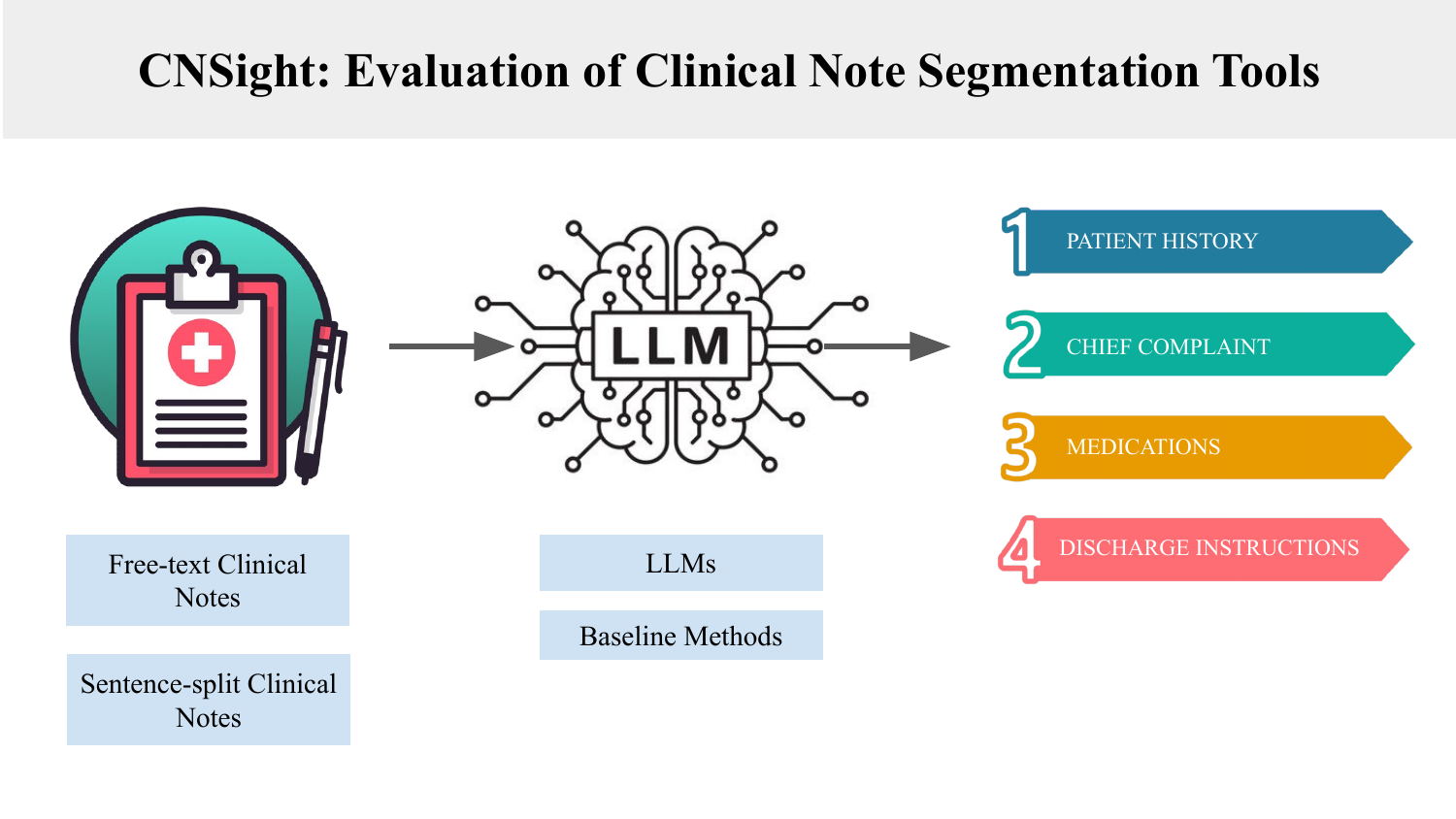}
    \caption{\textbf{LLMs enable automated segmentation of clinical notes.} Overview of the CNSight pipeline, where free-text or sentence-split clinical notes are processed by large language models and baseline methods to extract structured clinical sections such as patient history, chief complaint, medications, and discharge instructions.}
    \label{fig:your_image}
\end{figure*}

\section{Related Works}

Several prior studies have examined the problem of clinical note segmentation. Davis et al. (2025) introduced MedSlice, a pipeline that uses fine-tuned large language models from open source libraries to securely segment clinical notes \cite{davis2025medslice}. Their work highlights the effectiveness of LLMs in capturing section boundaries in a domain where notes often vary in length and structure.  

Earlier work by Ganesan and Subotin (2014) presented a supervised approach to clinical text segmentation. They evaluated the Essie 4 NLM library for its ability to identify and retrieve relevant sections from documents, demonstrating the utility of domain-specific tools for segmentation tasks \cite{ganesan2014supervised}.  
Edinger et al. (2018) evaluated multiple modeling approaches including regularized logistic regression, support vector machines, Naive Bayes, and conditional random fields for clinical text segmentation. Their study focused on improving cohort retrieval through accurate identification of note sections, emphasizing the importance of segmentation as a foundation for downstream clinical informatics applications \cite{edinger2018evaluation}.

\section{Methods}

We evaluate whether different modeling approaches can improve clinical note segmentation compared to naive baselines. 

\subsection{Datasets}

We use four main datasets in this project, derived from the MIMIC-IV corpus \cite{mimiciv} and supplemented with additional clinical-text resources. MIMIC-IV is a publicly available, de-identified dataset containing a wide range of clinical notes, including discharge summaries, physician notes, and semi-structured chart data. For evaluation across all datasets, we apply an 80/10/10 train/validation/test split using k-fold cross-validation. A distribution of the most frequent tags in these datasets are displayed in \autoref{app:dist}.

\paragraph{MIMIC Hospital.}  
From MIMIC-IV, we construct two custom datasets organized by note type using the labeled “Hospital Course” subset \cite{mimic_iv_ext_bhc_v1.2.0}:  
\begin{itemize}
    \item \textbf{MIMIC Hospital Sentences}: 1,000 unlabeled free-text clinical notes, split into 17,487 labeled clinical note sections segmented at the sentence level.  
    \item \textbf{MIMIC Hospital Freetext}: 1,000 unlabeled free-text clinical notes, generated from Mimic Hospital Sentences, which preserve the original narrative structure of the documentation.  
\end{itemize}

\paragraph{MIMIC Note Freetext.}  
This dataset consists of unlabeled free-text clinical notes drawn directly from MIMIC-IV \cite{mimic_iv_note_v2.2}. It provides a large, unstructured resource of clinical language for representation learning.

\paragraph{Augmented Clinical Notes.}  
This dataset consists of additional unlabeled free-text clinical notes curated for this project \cite{augmented_clinical_notes_agbonnet}. It is used to increase the diversity and size of the training corpus for representation learning.

\subsection{Models}

As baselines, we include a multinomial logistic regression classifier, a regex-based header matcher, and MedSpaCy \cite{eyre2021medspacy}, a clinical NLP toolkit for section detection and rule-based information extraction. These methods provide interpretable and lightweight points of comparison.  

For domain-specific transformer models, we evaluate LLaMA-2-7B \cite{touvron2023llama}, MedAlpaca-7B \cite{han2023medalpaca}, and Meditron-7B \cite{chen2023meditron}. These open-source models vary in their degree of biomedical adaptation and represent mid-scale transformer approaches tailored to clinical language tasks.  

Finally, we benchmark three API-based large language models: GPT-5-mini \cite{openai2025gpt5}, Gemini 2.5 Flash \cite{google2025gemini}, and Claude 4.5 Haiku \cite{anthropic2025claude}. These state-of-the-art systems offer broad-domain generalization and strong zero- and few-shot performance.

\section{Evaluation}

\begin{table*}[!t]
\centering
\setlength{\tabcolsep}{6pt}
\renewcommand{\arraystretch}{1.15}

\caption{\textbf{MIMIC Hospital results.} Models are evaluated on two datasets: \emph{MIMIC Hospital Sentences} (labeled clinical note sections) and \emph{MIMIC Hospital Freetext} (unlabeled free-text clinical notes). We report Precision, Recall, and F1-score for each dataset. Baseline 1 and Baseline 2 are included for comparison. We bold the highest performing model, and underline the second highest performing model per dataset. We additionally report the average F1 across tasks.}
\label{tab:overall-results}

\begin{tabular*}{\textwidth}{@{\extracolsep{\fill}} l ccc ccc c}
\toprule
\multirow{2}{*}{Model} & \multicolumn{3}{c}{MIMIC Hospital Sentences} & \multicolumn{3}{c}{MIMIC Hospital Freetext} & Avg. F1 \\
\cmidrule(lr){2-4} \cmidrule(lr){5-7} \cmidrule(lr){8-8}
 & Precision & Recall & F1 & Precision & Recall & F1 &  \\
\midrule
GPT-5-mini       & \textbf{85.4} & \textbf{81.6} & \textbf{80.8} & \underline{87.8} & \underline{50.2}  & \underline{63.9} & \underline{72.4} \\
Gemini 2.5 Flash & 81.3 & \underline{79.4} & \underline{78.5} & 87.3 & 46.6 & 60.8 & 69.7 \\
Claude 4.5 Haiku & \underline{83.3} & 72.9 & 76.5 & 82.9 & 33.6 & 47.8 & 62.2 \\
LLaMA-2-7B       & 17.9 & 12.4 & 8.9  & 87.3 & 37.5 & 52.4 & 30.7 \\
MedAlpaca-7B     & 5.5  & 3.1  & 0.6  & 83.0 & 25.2 & 38.6 & 19.6 \\
Meditron-7B      & 0.4 & 0.5 & 0.3 & 0.0 & 0.0 & 0.0 &  0.2 \\
Multinomial Logistic Regression & 79.1 & 77.7 & 74.3 & - & - & - & 74.3 \\
MedSpaCy         & 79.2 & 78.1 & 78.0 & \textbf{94.4} & \textbf{83.0} & \textbf{88.3} & \textbf{83.2} \\
\bottomrule
\end{tabular*}
\end{table*}


\subsection{Metrics}

Model performance is assessed using token-level Precision, Recall, and F1. For clinical note segmentation, we treat each predicted section boundary token as a classification decision. In this setup, a \textbf{True Positive (TP)} is a predicted boundary token that exactly matches a gold-standard boundary token, a \textbf{False Positive (FP)} is a predicted boundary token that does not correspond to any gold-standard boundary (i.e., a spurious split), and a \textbf{False Negative (FN)} is a gold-standard boundary token that the model fails to predict (i.e., a missed split).  

For the sentence classification task, we report \textbf{weighted F1}, which accounts for class imbalance by averaging per-class performance proportional to class frequency. This choice reflects the fact that some section types occur more frequently than others in the labeled dataset. For the freetext segmentation task, we instead report \textbf{micro-averaged F1}, which aggregates decisions across all boundaries. This metric provides a clearer measure of overall segmentation quality when sections are highly variable in length and frequency.  


\section{Results}

\begin{figure}[t]
    \centering
    \includegraphics[width=1.1\columnwidth]{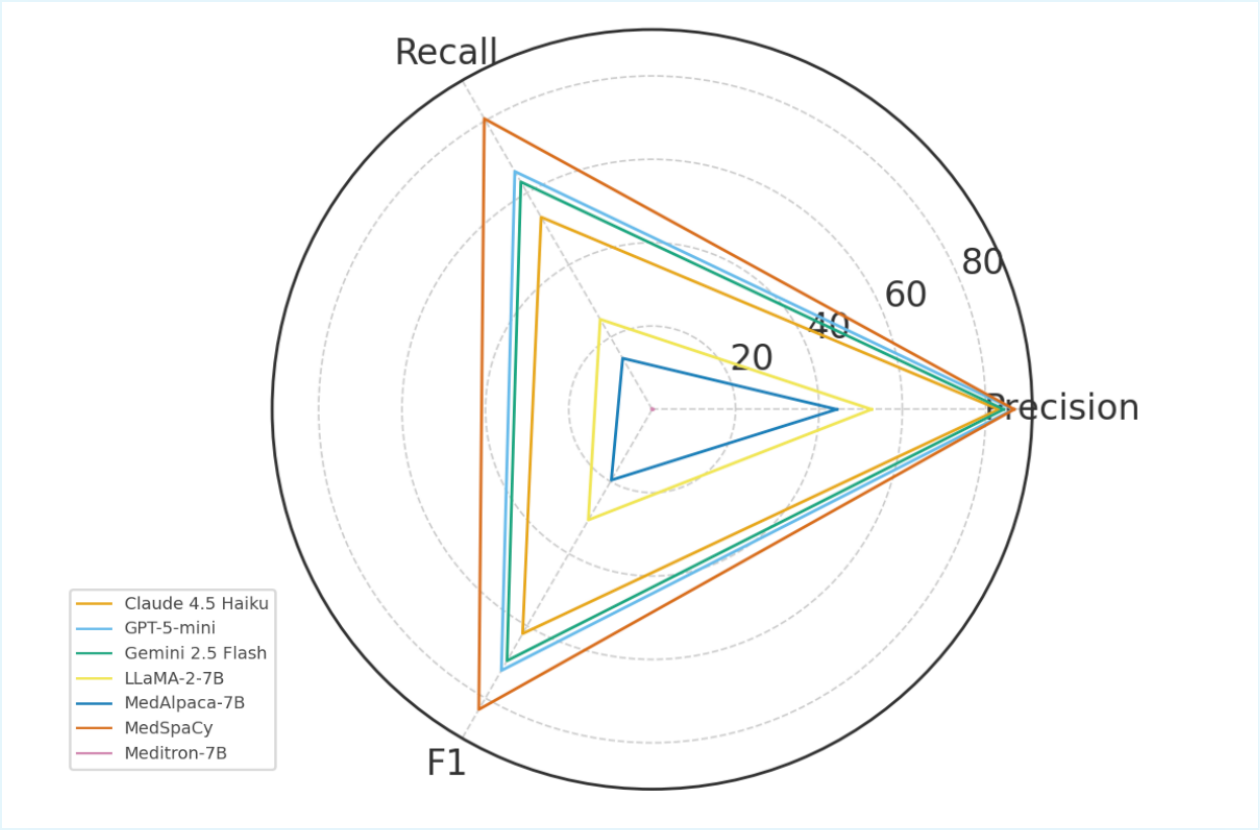}
    \caption{\textbf{Summary of metrics across models. } Comparison of precision, recall, and F1 scores across medical and general-purpose language models. General-purpose models outperform smaller domain-specific models.}
    \label{fig:radar}
\end{figure}

In \autoref{tab:overall-results}, we analyze three API-based LLM models, three HuggingFace locally hosted models, and two baselines for comparison. A summary of the model performance is observed in \autoref{fig:radar}.

\subsection{Baseline Analysis}
\paragraph{Traditional baselines achieve solid but limited performance on sentence-based analysis.} 
The Multinomial Logistic Regression classifier (Baseline 1) achieves strong performance on the sentence classification task, with a 74.3 F1. While this lags behind API-based LLMs by 4--6 points, it is still competitive with weaker commercial systems. 
MedSpaCy (Baseline 2) goes even further, reaching a 78.0 F1 on sentences---just 2.8 points below Gemini 2.5 Flash (78.5) and 2.8 points below Claude 4.5 Haiku (76.5). 
This demonstrates that rule-based, domain-specific NLP tools remain surprisingly strong in structured settings. 

\paragraph{Traditional baselines surpass LLMs with strong performance on freetext-based analysis.} 
MedSpaCy (Baseline 2) achieves the highest scores on freetext analysis, achieving a high score of 94.4 Precision, which is over 6 points higher than GPT-5-mini.

\subsection{API-Based LLMs}
\paragraph{Across the board, GPT-5-mini performed the best.} 
On both the Sentence Classification and Freetext tasks, GPT-5-mini achieved the highest scores across all metrics. 
Its F1 of 80.8 on sentences and 63.9 on freetext demonstrates not only high precision but also consistently strong recall, outperforming other commercial LLMs by several points. 
This indicates that GPT-5-mini is more robust at identifying relevant spans without overpredicting, balancing sensitivity with specificity in the clinical setting.  

\paragraph{Gemini 2.5 Flash and Claude 4.5 Haiku trail closely but unevenly.} 
Gemini performs competitively on both tasks, ranking second on freetext classification (60.8 F1) and maintaining strong performance on sentences (78.5 F1). 
Claude 4.5 Haiku exhibits high precision (83.3) but suffers from lower recall (72.9) on sentence classification, indicating a conservative prediction style. 
On freetext, its performance drops significantly (47.8 F1), underscoring difficulty in handling unstructured, noisy inputs.  

\subsection{Small and Domain-Specific LLMs}
\paragraph{Domain-tuned small LLMs underperform substantially.} 
Both LLaMA-2-7B and MedAlpaca-7B struggle, with sentence-level F1-scores of 8.9 and 0.6 respectively. 
This reflects their inability to capture the clinical context with limited scale and training. 
Although LLaMA-2-7B reaches moderate precision on freetext (87.3), its recall (37.5) lags, leading to mediocre overall performance (52.4 F1). 
Abbreviations and domain-specific shorthand often caused these models to miss boundaries, particularly in medication and lab result sections where formatting was dense.
Multi-line headers and irregular spacing also confused both rule-based baselines and smaller LLMs, leading to false positives where extraneous breaks were inserted.

\paragraph{Meditron-7B underperforms expectations.} 
Despite being designed as a biomedical-focused LLM, Meditron-7B fails to produce competitive results, with F1-scores of 0.3 on sentences and 0.0 on freetext. 
This suggests either significant mismatches between training and evaluation domains or limitations in model scale and optimization. 
Unlike MedAlpaca, which shows partial utility in freetext, Meditron’s outputs approach non-functional levels of performance.

\paragraph{ANOVA Testing.} We conducted a one-way ANOVA on model F1 scores across the two tasks. The test yielded $F = 2.45$ with $p = 0.18$, indicating that observed variations among models are not statistically significant at the 0.05 level. This suggests that while performance differences across tasks are visible, they should be interpreted with caution.  

\section{Human Evaluations}

To better understand how humans compare to automated systems on difficult sentence classification tasks, we conducted a small human evaluation using an unlabeled subset of clinical notes. Since it is challenging to obtain large expert-annotated corpora, our goal was not to create a new gold-standard dataset but to examine how general human annotators behave when presented with ambiguous or weakly structured inputs. This allows us to compare annotator behavior to the behavior of both baseline models and large language models and to evaluate where automated methods diverge from human judgment.

We sampled sentences from an unlabeled MIMIC-derived corpus by splitting raw notes into individual units. These sentences were then provided to nonexpert annotators who selected a section label from the same set of categories used in our supervised experiments. Because many extracted sentences contain partial headers, demographic fragments, or minimal context, this setup reflects a more challenging classification scenario than the segmented hospital course dataset used for model evaluation.

We report Cohen's Kappa and Percent Agreement between each system and the majority human label (\autoref{tab:agreement_results}). \textbf{Agreement between human annotators and API-based LLMs is moderate, with Claude and Gemini reaching Kappa values of 0.46 and agreements above 52 percent.} GPT 5 is slightly lower with 49.5 percent agreement. Baseline models show substantially weaker alignment with humans. Logistic regression reaches a Kappa of 0.19, while embedding-based and blank MedSpaCy variants range from 0.14 to 0.13.

\begin{table}[t]
\centering
\small
\setlength\tabcolsep{3pt}
\caption{Annotator Agreement: Humans vs. LLMs and Baselines on Clinical Section Tagging}
\label{tab:agreement_results}
\begin{tabular}{lrrr}
\hline
\textbf{Model} & \textbf{Kappa} & \textbf{Agreement} & \textbf{N} \\
\hline
\multicolumn{4}{c}{\textit{Braintrust LLM Models}} \\
\hline
Claude & \textbf{46.5} & \textbf{53.4}\% & 786 \\
Gemini & \underline{46.4} & \underline{52.6}\% & 1369 \\
GPT-5 & 43.2 & 49.5\% & 1578 \\
\hline
\multicolumn{4}{c}{\textit{Baseline Models}} \\
\hline
Log Regression   & 18.8 & 24.2\% & 3574 \\
Embedding Spacy  & 15.4 & 20.6\% & 3574 \\
Blank Spacy      & 13.6 & 18.2\% & 3574 \\
\hline

\end{tabular}
\end{table}

\begin{table*}[h]
\centering
\small
\setlength\tabcolsep{4pt}
\caption{Label Distribution Across Models for Ambiguous Header/Demographic Sections (First Row). Humans label 66\% of header rows as ``OTHER'', reflecting recognition of inherent ambiguity and multiple valid classification options. In contrast, baseline and LLM models exhibit strong, often inaccurate biases: Blank Spacy defaults to ALLERGIES (78/129), Embedding Spacy to FAMILY HISTORY (120/129), and Log Regression to ALLERGIES (108/129). This demonstrates that humans employ greater flexibility and judgment when faced with ambiguous sections, while automated systems tend toward fixed predictions.}
\label{tab:label_distribution}
\begin{tabular}{lrrrrrrr}
\hline
\textbf{Label} & \textbf{Human} & \textbf{Blank Spacy} & \textbf{Embedding Spacy} & \textbf{Log Reg} & \textbf{Claude} & \textbf{Gemini} & \textbf{GPT-5} \\
\hline
ALLERGIES & 45 & 78 & 0 & 108 & 12 & 15 & 45 \\
SEX & 108 & 0 & 0 & 0 & 9 & 27 & 3 \\
\textbf{OTHER} & \textbf{189} & 0 & 0 & 0 & 0 & 0 & 0 \\
SERVICE & 27 & 0 & 0 & 0 & 0 & 0 & 0 \\
HISTORY OF PRESENT ILLNESS & 9 & 3 & 0 & 6 & 0 & 6 & 9 \\
MAJOR SURGICAL OR INV. PROC. & 9 & 0 & 0 & 0 & 0 & 0 & 0 \\
FAMILY HISTORY & 0 & 0 & 120 & 0 & 0 & 0 & 0 \\
DISCHARGE DIAGNOSIS & 0 & 39 & 0 & 0 & 0 & 0 & 0 \\
PHYSICAL EXAM & 0 & 0 & 9 & 0 & 0 & 0 & 0 \\
PERTINENT RESULTS & 0 & 9 & 0 & 9 & 0 & 0 & 0 \\
ATTENDING & 0 & 0 & 0 & 0 & 3 & 0 & 0 \\
CHIEF COMPLAINT & 0 & 0 & 0 & 3 & 3 & 0 & 0 \\
PAST MEDICAL HISTORY & 0 & 0 & 0 & 3 & 0 & 0 & 0 \\
\hline
\textbf{Total} & \textbf{387} & \textbf{129} & \textbf{129} & \textbf{129} & \textbf{27} & \textbf{48} & \textbf{57} \\
\hline
\end{tabular}
\end{table*}

To further understand disagreement patterns, we analyzed the labels assigned to the most ambiguous subset of sentences, which primarily consisted of header-like fragments or demographic rows. As shown in \autoref{tab:label_distribution}, humans label 66 percent of these sentences as OTHER, recognizing that many of these items have no single correct section label. Automated systems behave very differently. The logistic regression model strongly defaults to ALLERGIES, embedding-based MedSpaCy predicts FAMILY HISTORY for nearly all items, and blank MedSpaCy heavily favors ALLERGIES. API-based LLMs also exhibit biases but with less extreme concentration in a single class.

These results indicate that human annotators demonstrate flexibility when dealing with ambiguous clinical text and do not force a strict interpretation of section categories. Automated models, in contrast, tend to collapse uncertainty into a small set of high-probability labels. This evaluation was therefore used not as ground truth for training or benchmarking but as a way to measure differences in behavior between humans and automated systems when faced with difficult and underspecified sentences.

\section{Conclusion}

In this study, we evaluated a wide range of clinical note segmentation approaches, including rule-based systems, traditional classifiers, domain-specific transformer models, and state-of-the-art large language models. Our results show that general lightweight baselines like MedSpacy remain strong on unstructured freetext tasks, but their performance drops when applied to structured sentences. API-based large language models achieve the highest overall performance on Sentence-level evaluations, with GPT 5 providing the most consistent gains across both sentence classification and freetext segmentation.

The human evaluation highlights an additional dimension of this problem. When presented with difficult or weakly contextualized clinical sentences, human annotators show flexible judgment and frequently select ambiguous categories rather than forcing a narrow interpretation. In contrast, automated systems tend to collapse uncertainty into a small set of high-probability labels. This pattern is especially visible in header and demographic fragments, where baselines and LLMs often apply strong but inaccurate defaults. These findings indicate that segmentation models are sensitive not only to domain structure but also to the way ambiguous input is represented and labeled.

Overall, our study demonstrates that high quality segmentation of clinical notes depends on both model capability and the nature of the input text. Large models offer strong generalization to structured documentation, while custom rule-based and classical methods remain valuable for un-structured formats. The human evaluation further underscores the importance of understanding annotation behavior when designing or interpreting segmentation systems. Future work should explore methods that better capture uncertainty in ambiguous sections and that leverage human-like flexibility in classification decisions.

\bibliography{custom}

\appendix



\section{Acknowledgments}
The authors gratefully acknowledge Braintrust for providing access to API-based models and for their assistance with system setup and integration, which facilitated the execution of these experiments.

\section{Data Distribution}
\label{app:dist}

\begin{figure*}[t]
    \centering
    \includegraphics[width=0.9\textwidth]{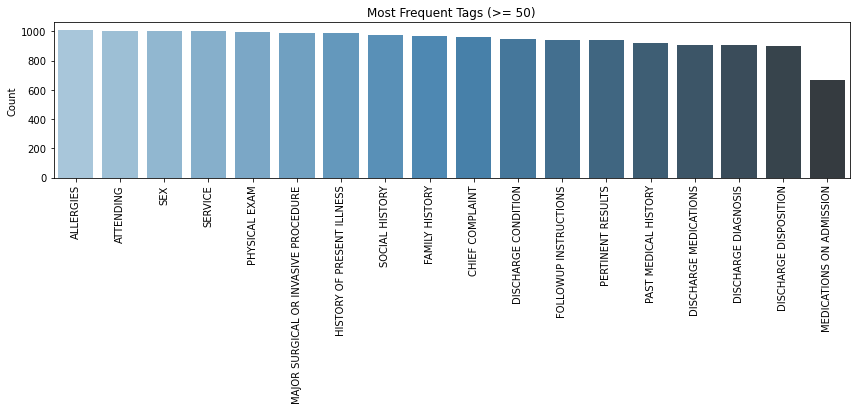}
    \caption{\textbf{A small set of clinical tags dominates the dataset.} Bar chart showing the most frequent clinical tags (occurring at least 50 times), with counts on the y-axis and tag categories on the x-axis.}
    \label{fig:dist}
\end{figure*}

\begin{figure*}[t]
    \centering
    \includegraphics[width=0.9\textwidth]{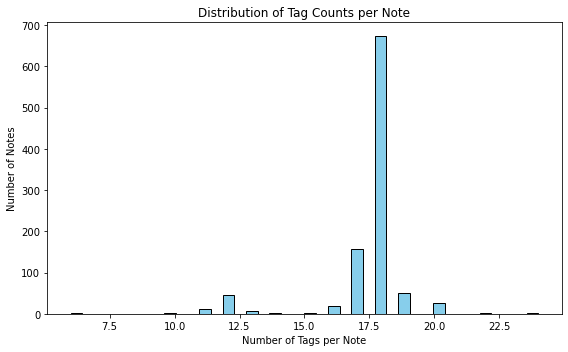}
    \caption{\textbf{Most notes contained approximately 18 tags.} Bar chart showing the distribution of the number of tags across all notes, shows a relatively consistent number across the dataset.}
    \label{fig:dist}
\end{figure*}

\begin{figure*}[t]
    \centering
    \includegraphics[width=0.9\textwidth]{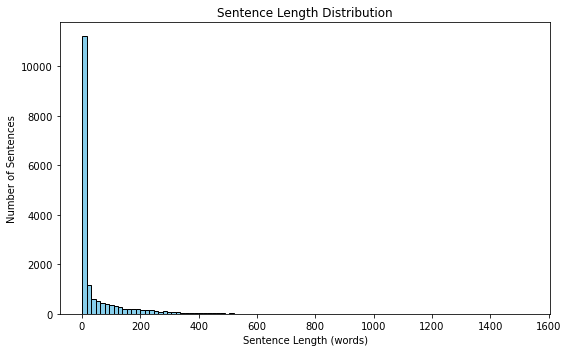}
    \caption{\textbf{Sentence length variation.} Bar chart showing the sentence length distributions across tags, shows a high number of short "sentences", referring to phrases and small extracted header sections for increased precision.}
    \label{fig:dist}
\end{figure*}

\end{document}